\documentclass[11pt,letterpaper,logo]{seu-paisys}

\usepackage[utf8]{inputenc}
\usepackage[T1]{fontenc}
\usepackage{hyperref}
\usepackage{url}
\usepackage{booktabs}
\usepackage{amsfonts}
\usepackage{nicefrac}
\usepackage{microtype}
\usepackage{xcolor}
\usepackage{graphicx}

\usepackage{multirow}
\usepackage{amsmath}
\usepackage{tabularx}
\usepackage{array}
\usepackage{CJKutf8}
\usepackage{xspace}

\newcommand{\system}{\texttt{Embodied.cpp}\xspace}

\usepackage{tikz}
\usetikzlibrary{arrows.meta}

\newcommand{\llamacpp}{llama.cpp}
\newcommand{\vlacpp}{vla.cpp}
\newcommand{\cmark}{\ding{51}}
\newcommand{\xmark}{\ding{55}}
\newcommand{\omark}{\ensuremath{\circ}}
\newcommand{\tmark}{\ensuremath{\triangle}}
\newcommand{\seumark}{\includegraphics[height=1.2em]{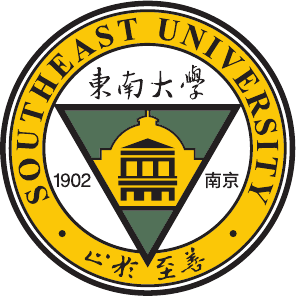}}
\newcommand{\njumark}{\includegraphics[height=1.2em]{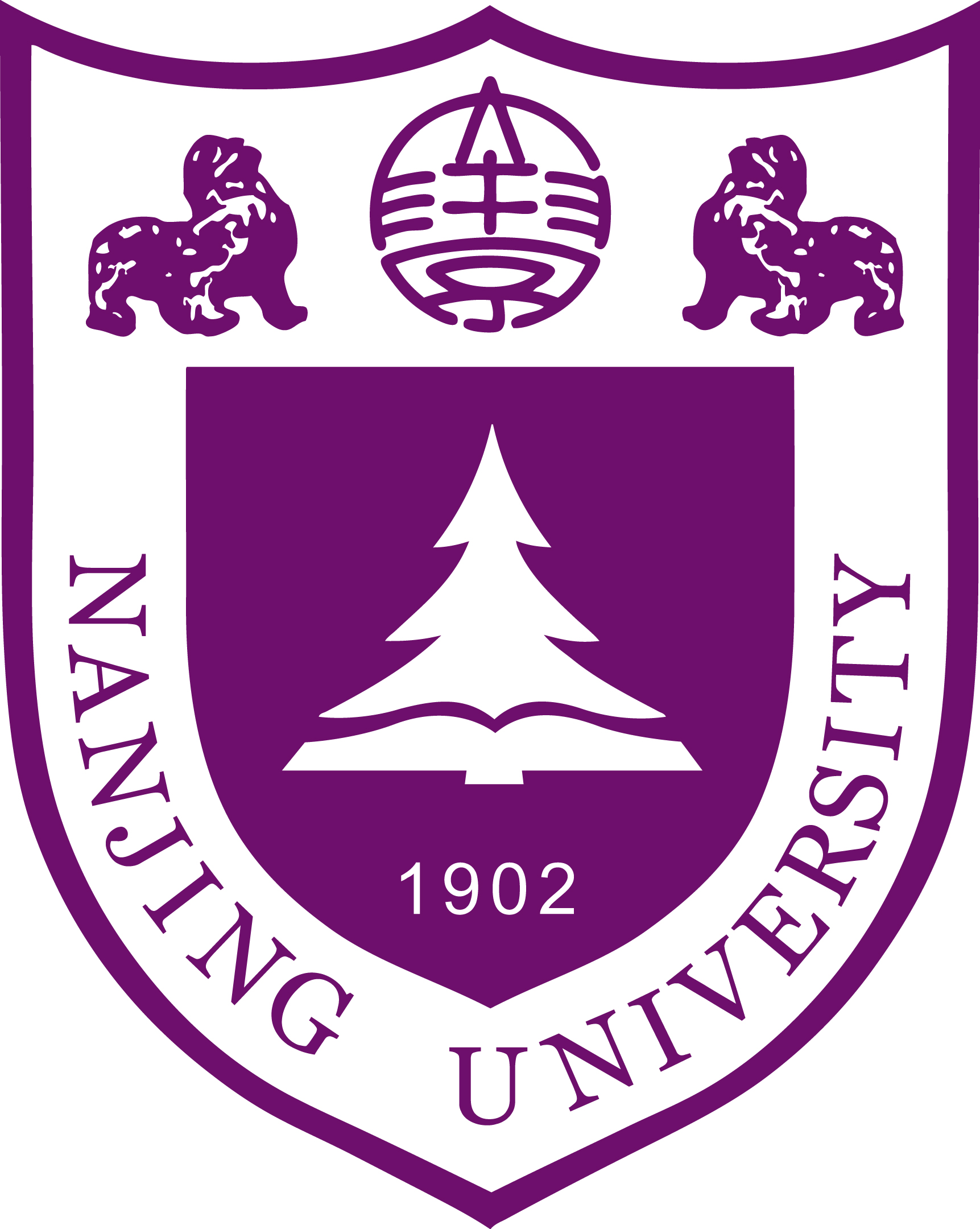}}
\newcommand{\thumark}{\includegraphics[height=1.2em]{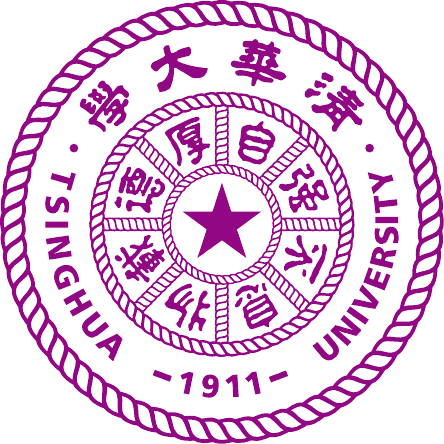}}
\newcommand{\msmark}{\includegraphics[height=1.2em]{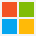}}

\renewcommand{\today}{2026-07-01}
\title{\system: A Portable Inference Runtime of Embodied AI Models on Heterogeneous Robots}

\author{%
  {\large\textbf{Ling Xu}$^{1}$ \quad \textbf{Borui Li}$^{1,\dagger}$ \quad \textbf{Hao Wu}$^{2,\dagger}$ \quad
    \textbf{Chuyu Han}$^{2}$ \quad \textbf{Xiangyu Li}$^{4}$  \quad \textbf{Mohan Hua}$^{2}$ \quad
    \textbf{Shiqi Jiang}$^{3}$ \quad
  \textbf{Ting Cao}$^{4}$ \quad \textbf{Chuanyou Li}$^{1}$ \quad \textbf{Sheng Zhong}$^{2}$ \quad \textbf{Shuai Wang}$^{1}$\\}
  \vspace{0.2em}
  {\normalsize
    $^{1}$\ \seumark\ Southeast University \qquad
    $^{2}$\ \njumark\ Nanjing University \qquad
    $^{3}$\ \msmark\ Microsoft Research \qquad \qquad \qquad
  $^{4}$\ \thumark\ Institute for AI Industry Research (AIR), Tsinghua University }

  {\normalsize $^{\dagger}$Project Leader}
}

\begin{document}

%\begin{abstract}
%\end{abstract}

\begin{abstract}
  \quad Embodied AI models now span vision-language-action (VLA) models and world-action models (WAMs), but practical deployment remains fragmented across model-specific Python stacks, backend assumptions, and robot-side glue code, especially on heterogeneous edge devices. Existing inference runtimes are designed mainly for request-response serving and therefore do not satisfy the runtime contract of embodied deployment: multi-rate execution inside closed-loop control, latency-first batch-1 inference on heterogeneous hardware, and extensible embodied interfaces beyond fixed token I/O. We present \textbf{\system}, a portable C++ inference runtime for embodied models. Based on an architectural analysis of representative VLA models and WAMs, \system captures a shared execution path and organizes it into five layers: input adapters, sequence builders, backbone execution, head plugins, and deployment adapters. The runtime provides modular multi-rate execution, latency-first fused inference, and extensible operator and I/O support, enabling deployment across heterogeneous devices, robots, and simulators through one backend abstraction. We evaluate \system on three VLA and two WAM models, using normalized comparisons across Python and C++ quantization configurations. Overall, \system achieves $1.05\times$--$2.70\times$ inference speedups and 7\%--77\% lower VRAM relative to Python baselines, while maintaining near-baseline success for most configurations. These results show that \system improves deployment efficiency while preserving high control quality across diverse embodied model architectures.

  \quad Project Link: \href{https://github.com/SEU-PAISys/Embodied.cpp}{https://github.com/SEU-PAISys/Embodied.cpp}
\end{abstract}

\maketitle

%\section{Introduction}
%
%Embodied AI is advancing quickly, and its practical usage is increasingly shaped by a diverse set of model paradigms. Recent progress now spans vision-language-action (VLA) models and world-action models (WAMs). At the same time, the surrounding ecosystem for data, training, and simulation has become substantially stronger, supported by projects such as LeRobot, OpenVLA, Open X-Embodiment, ManiSkill, LIBERO, and Isaac Sim~\cite{lerobot,openvla,openx,maniskill,libero,isaacsim}. The central question is therefore no longer whether embodied models can be trained, but how an evolving family of such models can be turned into reliable robot-side systems.

\section{Introduction}

Academia and industry have already produced a rapidly growing set of embodied models. Recent systems now span vision-language-action (VLA) models and world-action models (WAMs), showing substantial progress in model architectures and capabilities~\cite{openvla,pi0,pi05,grootn1,lingbotva,wam-survey,li2026stop}. At the same time, the surrounding ecosystem for data, training, and simulation has also improved substantially through projects such as LeRobot, Open X-Embodiment, ManiSkill, LIBERO, and Isaac Sim~\cite{lerobot,openx,maniskill,libero,isaacsim}. But practical impact does not end at model construction. To become reliable robot-side systems, these models must run on heterogeneous and resource-constrained devices, from Jetson and RK-based boards to x86 edge boxes and workstation-class robots, without rebuilding the software stack for each new model family.

%That question is fundamentally a deployment question. Unlike cloud-oriented language or vision-language serving, embodied inference lives inside a control loop: observations arrive continuously, actions must be produced under timing constraints, and failures are expressed in physical behavior rather than degraded user experience. For robots, efficient on-device deployment is especially important. Robots keep moving while sensing and acting, and many tasks require high-precision, low-latency action generation. In practice, this means embodied models need to run locally on Jetson devices, RK-based boards, x86 edge boxes, or workstation-class robots with predictable latency and a manageable software stack.

Achieving this requires a unified inference runtime. Existing inference runtimes, however, do not yet provide such support. General LLM or VLM runtimes are designed for request-response serving, relatively uniform token interfaces, and throughput-oriented optimization~\cite{llama-cpp,onnxruntime,sglang,vllm-omni}. Embodied inference, by contrast, lives inside a closed-loop interaction process with robot- and simulator-side dependencies. As a result, even a strong checkpoint must be stitched into Python research code, backend-specific inference paths, handwritten sensor wrappers, and platform-specific control logic before it can act on a robot. As embodied architectures diversify, this integration burden only grows, making a portable inference runtime for embodied AI models increasingly necessary.

%Current deployment practice does not yet meet that need. Even when a strong checkpoint is available, real deployment is still fragmented across Python-heavy research code, backend-specific inference paths, hand-written sensor wrappers, and platform-specific control loops. This makes embodied deployment a recurring integration effort rather than a reusable systems layer. The problem becomes even sharper as paradigms diversify: each new model family seems to arrive with a slightly different execution story, encouraging teams to rebuild the same infrastructure again and again.

The challenge arises because embodied deployment fundamentally changes the runtime contract. Compared with conventional LLM or VLM serving, an embodied inference runtime must satisfy three key requirements. \textbf{(1) Multi-rate execution.} Embodied inference is not a single synchronous forward pass: perception encoders, transformer backbones, predictive branches, and action heads may need to run at different rates inside the same control loop, as increasingly visible in hierarchical and fast-slow VLA systems as well as recent WAMs~\cite{pi05,grootn1,lingbotva}. \textbf{(2) Latency-first closed-loop control.} The optimization target is not throughput but stable closed-loop control, which requires low latency, low jitter, and efficient batch-1 execution on heterogeneous edge hardware. \textbf{(3) Extensible embodied interfaces.} Embodied deployment must go beyond fixed token I/O and absorb custom operators, embodied multimodal inputs, and heterogeneous outputs ranging from action chunks to predicted futures. These are not peripheral engineering details; together they define the runtime contract that a practical embodied inference runtime must satisfy.

%This report argues that the right response is a dedicated runtime substrate for embodied deployment. We position \system as a portable C++ runtime that provides a unified execution layer for embodied models across heterogeneous edge devices and robots. The choice of C++ and portability is therefore a systems decision, not a branding choice: a portable C++ runtime offers a small dependency surface, direct control over memory and scheduling, predictable latency, and a practical path to deployment across different robot-side devices. Our core claim is that, despite rapid model evolution, recent embodied systems already expose enough architectural regularity to support such a stable runtime abstraction.

Motivated by these observations, we present \system, a portable C++ inference runtime for embodied models. Its design directly addresses the three challenges: modular multi-rate execution decouples components with different refresh frequencies; latency-first fused execution targets predictable, small-batch control on heterogeneous devices; and extensible operator and embodied I/O support turns model-specific heads, interfaces, and backends into pluggable runtime modules rather than one-off glue code. These principles are realized in a unified five-layer architecture spanning input adapters, sequence builders, backbone execution, head plugins, and deployment adapters. The result is an inference runtime with a stable, reusable core, while still leaving enough extension surface for new embodied model variants.

% The core bet is that, despite rapid model evolution, recent embodied systems already expose enough architectural regularity to support a stable runtime abstraction. We make that case with the architecture in Section~2 and the VLA deployment results in Section~3.

We make three contributions:
\begin{itemize}
  \item An architectural analysis of representative embodied models spanning vision-language-action (VLA) models and world-action models (WAMs), revealing a shared execution path whose main divergences are confined to a small set of pluggable heads and predictive modules.
  \item The design of \system as a portable C++ inference runtime that maps this shared path into five layers: input adapters, sequence builders, backbone execution, head plugins, and deployment adapters, with backend abstraction for heterogeneous edge devices.
  \item We evaluate \system on three VLA models and two representative WAM deployments, showing improved efficiency while maintaining high accuracy.
\end{itemize}

\section{Related Work and Motivation}
\label{sec:related_work}

\subsection{Embodied AI Model and Architectural Analysis}

\begin{figure*}[t]
  \centering
  \includegraphics[width=\textwidth]{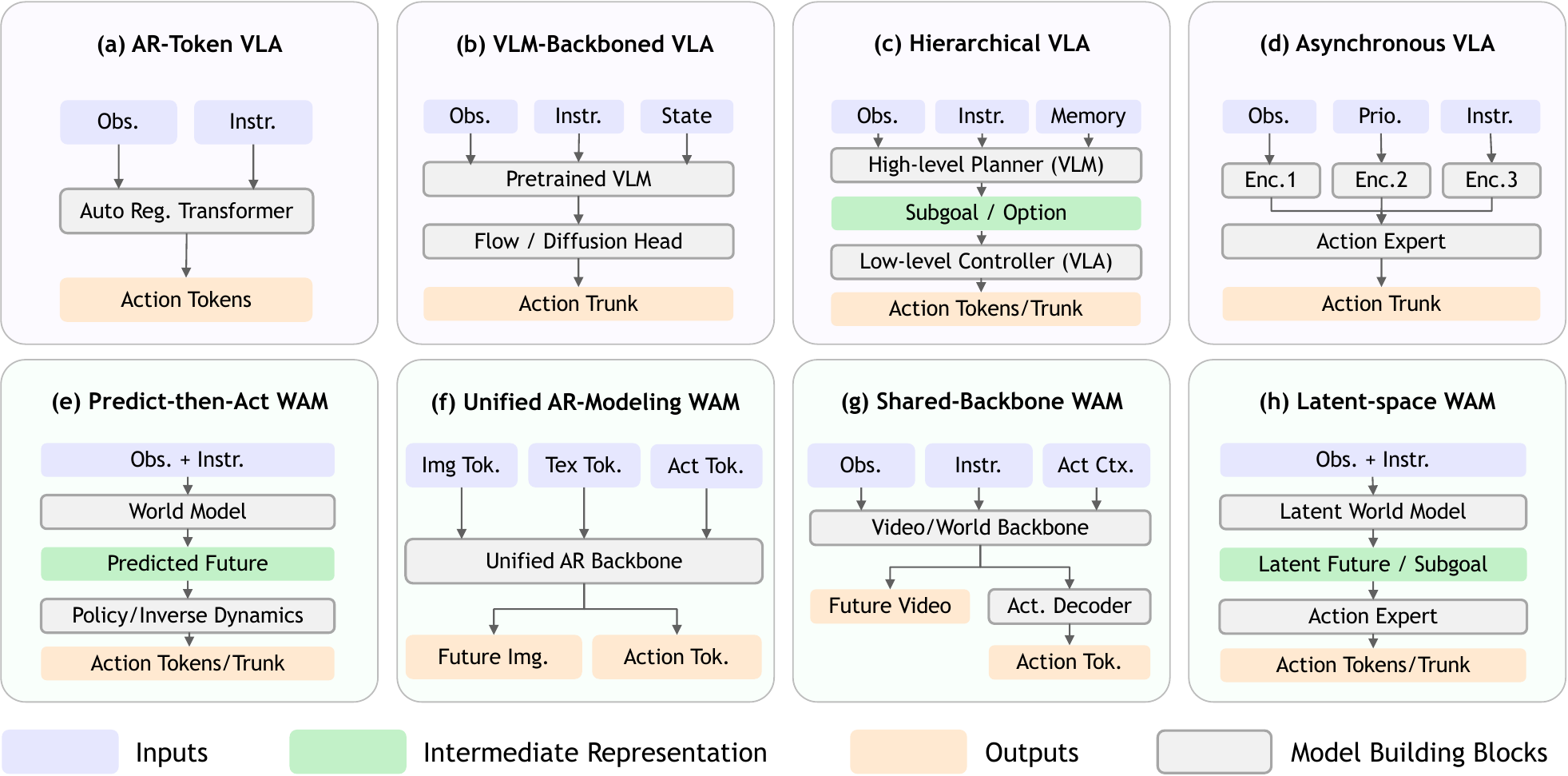}
  \caption{Architectural taxonomy of embodied models. We first separate VLA models (a-d) from WAMs (e-h), and then organize each family by internal inference structure.}
  \label{fig:architecture-taxonomy}
\end{figure*}

From an inference-runtime perspective, recent embodied models can be organized into two online families: vision-language-action (VLA) models and world-action models (WAMs). As summarized in Figure~\ref{fig:architecture-taxonomy} and Table~\ref{tab:architecture-serving}, VLA models primarily realize a perception-to-action path, while WAMs make future prediction an explicit part of online control~\cite{openvla,pi0,wam-survey}. This split matters directly to systems design because it determines whether the runtime mainly executes an action policy or must jointly maintain predictive state and action generation.

Within the VLA family, the architectural progression is from monolithic action generation to increasingly structured modular execution. AR-Token VLA models such as RT-2 and OpenVLA use one shared backbone to autoregressively generate action tokens~\cite{rt2,openvla}. VLM-Backboned VLA models such as Octo, pi0, pi0.5, and MuseVLA still rely on a strong shared visual-language backbone, but pair it with more explicit continuous-action heads~\cite{octo,pi0,pi05,musevla}. More recent systems then branch into two modular directions. Hierarchical VLA models split semantic planning from low-level control, with a planner producing subgoals for a downstream controller, as in Hi Robot, GeneralVLA, RT-H, and Gemini Robotics 1.5~\cite{hirobot,generalvla,rth,geminirobotics15}. Asynchronous VLA models instead split execution by time scale: decoupled modules run at different rates and coordinate through buffered state, as in GR00T N1, Fast-in-Slow, and DAM-VLA~\cite{grootn1,fis,damvla}. The key runtime takeaway is that even within VLA models, the execution unit is no longer uniformly a single synchronous forward pass.

The WAM family is organized by how future prediction is coupled with action generation. Predict-then-Act WAMs such as UniPi keep the two stages explicit: a world model predicts future states, and a downstream action expert consumes them~\cite{unipi}. Unified AR-Modeling WAMs collapse future modeling and action generation into one autoregressive token space, as in WorldVLA and LingBot-VA~\cite{worldvla,lingbotva}. Shared-Backbone WAMs reuse one backbone across world modeling and action generation, while still exposing separate blocks that can potentially be scheduled or extended independently, as in DreamZero, Fast-WAM, Cosmos Policy, and the Unified Video Action Model (UWM)~\cite{dreamzero,fastwam,cosmospolicy,uwm}. Latent-space WAMs compress the predictive path further by generating a compact latent future or subgoal for a downstream action expert, as in LaWAM and Being-H0.7~\cite{lawam,beingh07}. Compared with VLA models, WAMs therefore broaden the runtime contract from direct action decoding to coordinated management of predictive computation and control computation.

\begin{table*}[t]
  \centering
  \caption{Architecture-oriented comparison of embodied AI models.}
  \label{tab:architecture-serving}
  \footnotesize
  \setlength{\tabcolsep}{3.8pt}
  \renewcommand{\arraystretch}{1.14}
  \begin{tabularx}{\textwidth}{>{\raggedright\arraybackslash}p{0.22\textwidth} >{\raggedright\arraybackslash}p{0.31\textwidth} >{\centering\arraybackslash}p{0.09\textwidth} X}
    \toprule
    \rowcolor{black!5}
    \textbf{Subtype} & \textbf{Architectural Characteristic} & \textbf{Structure} & \textbf{Representative Models} \\
    \midrule
    \rowcolor{black!2}
    \multicolumn{4}{l}{\textbf{Vision-Language-Action (VLA) Models}} \\
    AR-Token VLA & One backbone autoregressively generates action tokens. & Monolithic & OpenVLA~\cite{openvla}, RT-2~\cite{rt2} \\
    VLM-Backboned VLA & A shared pretrained VLM feeds a continuous action head. & Modular & Octo~\cite{octo}, pi0~\cite{pi0}, pi0.5~\cite{pi05}, MuseVLA~\cite{musevla} \\
    Hierarchical VLA & A high-level planner produces subgoals for a low-level controller. & Modular & Hi Robot~\cite{hirobot}, GeneralVLA~\cite{generalvla}, RT-H~\cite{rth}, Gemini Robotics 1.5~\cite{geminirobotics15} \\
    Asynchronous VLA & Decoupled modules run at different rates through buffered coordination. & Modular & GR00T N1~\cite{grootn1}, Fast-in-Slow~\cite{fis}, DAM-VLA~\cite{damvla} \\
    \addlinespace[2pt]
    \rowcolor{black!2}
    \multicolumn{4}{l}{\textbf{World-Action Models (WAM)}} \\
    Predict-then-Act WAM & An explicit world model predicts future states before a downstream action expert acts. & Modular & UniPi~\cite{unipi} \\
    Unified AR-Modeling WAM & Future world and robotic action are jointly generated in one autoregressive token space. & Monolithic & WorldVLA~\cite{worldvla}, LingBot-VA~\cite{lingbotva} \\
    Shared-Backbone WAM & World modeling and action generation share one backbone, while auxiliary blocks can run at different rates. & Modular & DreamZero~\cite{dreamzero}, Fast-WAM~\cite{fastwam}, Cosmos Policy~\cite{cosmospolicy}, UWM~\cite{uwm} \\
    Latent-space WAM & A compact latent future or subgoal is predicted and then consumed by an action expert. & Modular & LaWAM~\cite{lawam}, Being-H0.7~\cite{beingh07} \\
    \bottomrule
  \end{tabularx}
\end{table*}

Taken together, Figure~\ref{fig:architecture-taxonomy} and Table~\ref{tab:architecture-serving} suggest three implications for inference runtime design. First, structural organization is now a first-class model property: some systems remain monolithic, but many recent ones are better understood as interacting planner, backbone, world-model, and action modules. Second, intermediate states such as subgoals, buffered context, predicted futures, and latent futures have become explicit runtime objects rather than hidden internal details. Third, timing structure is increasingly model-defined, especially in hierarchical and asynchronous systems, so a practical embodied inference runtime must support stateful multi-component orchestration instead of assuming one uniform synchronous request-response path.

\subsection{Existing Inference Runtimes for AI Models}

AI model inference has been well-studied in the past few years~\cite{wu2021pecam,wu2020emo,zeng2025h2o,li2024agent,li2025infscaler,li2025mobilora}.
General-purpose inference runtimes already cover much of the shared execution substrate. \llamacpp{} emphasizes lightweight C/C++ local inference, broad hardware coverage, and practical packaging~\cite{llama-cpp}; ONNX Runtime provides cross-platform acceleration through multiple execution providers~\cite{onnxruntime}; and serving systems such as SGLang and vLLM-Omni target large-scale LLM/VLM inference with richer multimodal pipelines~\cite{sglang,vllm-omni}. These systems are strong inference runtimes, but they primarily target request-response workloads with relatively uniform interfaces.

Embodied settings impose a different contract: closed-loop control, heterogeneous inputs and outputs, persistent state, batch-1 latency sensitivity, and direct robot or simulator integration. Simulation platforms such as ManiSkill, LIBERO, and Isaac Sim are important for training and benchmarking~\cite{maniskill,libero,isaacsim}, but they are not themselves embodied inference runtimes. The closest recent step is \vlacpp{}, which brings seven VLA architectures into one portable C++ inference runtime and reports execution from consumer GPUs to embedded devices together with an on-robot stress test~\cite{vla-cpp}. This is an important milestone, but it remains VLA-centric.

Table~\ref{tab:runtime-compare} makes the gap concrete from two complementary angles: model-family coverage and deployment-facing capability. The first question is whether a runtime directly supports VLA models or WAMs, and whether it provides explicit optimization support for modular multi-component models rather than only monolithic inference graphs. The second is whether it can actually carry those models to embodied deployment, including edge execution, heterogeneous-device cooperation, and direct connection to robots or simulators. Existing systems are either generic inference runtimes without first-class support for embodied models, or embodied inference runtimes specialized to a single model family.

\begin{table*}[t]
  \centering
  \caption{Comparison of inference runtimes from the perspective of embodied model coverage and deployment readiness. `VLA' and `WAM' denote first-class support for the corresponding embodied model families. `Modular' denotes support or optimization for multi-component embodied models rather than only monolithic inference graphs. `Edge' denotes whether the runtime can run with deployment-oriented optimization on edge or end-side devices. `Hetero. HW' denotes whether one runtime can jointly use heterogeneous devices rather than only switch among backends. \cmark: native support. \xmark: unsupported. \omark: achievable only with substantial custom integration. \tmark: limited or partial support.}
  \label{tab:runtime-compare}
  \footnotesize
  \setlength{\tabcolsep}{4.1pt}
  \renewcommand{\arraystretch}{1.14}
  \begin{tabularx}{\textwidth}{>{\raggedright\arraybackslash}X >{\centering\arraybackslash}p{0.07\textwidth} >{\centering\arraybackslash}p{0.07\textwidth} >{\centering\arraybackslash}p{0.09\textwidth} >{\centering\arraybackslash}p{0.07\textwidth} >{\centering\arraybackslash}p{0.11\textwidth} >{\centering\arraybackslash}p{0.08\textwidth} >{\centering\arraybackslash}p{0.10\textwidth}}
    \toprule
    \rowcolor{black!5}
    \textbf{System} & \textbf{VLA} & \textbf{WAM} & \textbf{Modular} & \textbf{Edge} & \textbf{Hetero. HW} & \textbf{Robot} & \textbf{Simulator} \\
    \midrule
    \llamacpp{}~\cite{llama-cpp} & \xmark & \xmark & \xmark & \cmark & \tmark & \xmark & \xmark \\
    \mbox{ONNX Runtime}~\cite{onnxruntime} & \omark & \omark & \omark & \cmark & \cmark & \xmark & \xmark \\
    \textsc{SGLang}~\cite{sglang} & \xmark & \xmark & \xmark & \xmark & \xmark & \xmark & \xmark \\
    \textsc{vLLM-Omni}~\cite{vllm-omni} & \omark & \omark & \tmark & \xmark & \tmark & \xmark & \xmark \\
    \vlacpp{}~\cite{vla-cpp} & \cmark & \xmark & \tmark & \cmark & \tmark & \cmark & \tmark \\
    \system{} (ours) & \cmark & \cmark & \cmark & \cmark & \cmark & \cmark & \cmark \\
    \bottomrule
  \end{tabularx}
\end{table*}

%The main gap is therefore not raw inference alone. What remains underbuilt is a deployment substrate that simultaneously supports multiple embodied paradigms, portable edge execution, embodied-aware packaging, and direct integration with robot-side control loops. \system is designed to occupy that systems layer.

Overall, existing systems still fall short of first-class coverage of embodied models, modular model optimization, and direct deployment interfaces. What remains underbuilt is an embodied inference runtime that jointly supports VLA and WAM execution and runs on edge devices, heterogeneous hardware, robots, and simulators.

\subsection{Existing Acceleration Methods for Model Inference}

In this section, we introduce existing acceleration methods for resource-constrained devices and embodied models.

\textbf{Accelerating inference on resource-constrained devices.}
% Early work on resource-constrained inference established that deployment efficiency is not obtained solely by selecting a faster processor: the computation graph, the intermediate representation, and the execution boundary must be optimized together. Sparsifying and separating DNN layers made it possible to trade local computation and transmitted activation size on wearables~\cite{bhattacharya2016sparsification}. Dynamic DNN surgery~\cite{hu2019surgery} subsequently adapted the executed network structure to changing edge resources, while deep compressive offloading~\cite{yao2020compressive} jointly selected an offloading boundary and compressed the intermediate feature sent across it. Building on this view of inference as a distributed computation graph, CoEdge~\cite{zeng2021coedge} and DistrEdge~\cite{hou2022distredge} adaptively partitioned work across heterogeneous edge devices, and Autodidactic Neurosurgeon~\cite{zhang2021autodidactic} used online feedback to refine collaborative inference decisions under changing mobile-edge conditions.
Existing works explicitly consider the heterogeneity inside a resource-constrained edge device.
NN-Meter~\cite{zhang2021nnmeter} showed why FLOPs alone cannot predict deployment latency and constructed device-specific latency models for diverse edge targets. CoDL~\cite{jia2022codl} and Band~\cite{jeong2022band} coordinated CPU--GPU or broader mobile-processor co-execution for one or multiple DNNs, while EdgeNN~\cite{zhang2023edgenn} exploited CPU--GPU-integrated edge architectures to reduce inefficient execution and movement between the two processors. At the network boundary, AoDNN~\cite{huang2022aodnn} automatically selected mobile-Web offloading decisions, CoActo~\cite{bin2024coacto} enabled fine-grained concurrent local--edge execution, and Jellyfish~\cite{nigade2022jellyfish} scheduled inference services against time-varying edge-network conditions. Together, these efforts move beyond static device placement toward performance-model-driven partitioning, concurrent execution, and network-aware scheduling.

Besides heterogeneity, resource constraints on edge devices make resource management a first-class concern~\cite{xiao2026surviving}. Niagara~\cite{xu2023niagara} schedules DNN services over heterogeneous edge processors; SensiX++~\cite{min2023sensix} brings multi-tenant model serving and lifecycle management to sensory edge devices; and Pantheon~\cite{han2024pantheon} and Miriam~\cite{zhao2023miriam} address priority-sensitive multi-DNN execution on edge GPUs through preemption and elastic-kernel coordination, respectively. InfScaler~\cite{li2025infscaler} further performs asymmetric auto-scaling across multiple non-identical accelerators. Complementary work reduces the device-local cost of each execution: FlexNN~\cite{li2024flexnn} and SwapNet~\cite{wang2024swapnet} manage DNN inference within tight and dynamic memory budgets; vMCU~\cite{zheng2024vmcu} co-designs memory management and kernels for microcontrollers; and Boosting DNN Cold Inference~\cite{yi2023boosting} targets the otherwise neglected first-inference path.

Recent advances in large language models further complicate the problem of accelerating models on resource-constrained devices. LLM in a Flash~\cite{alizadeh2024flash} extends feasible model size with flash-resident, on-demand weights; HeteGen~\cite{zhao2024hetegen} overlaps CPU--GPU computation and data movement for resource-constrained LLM inference; and EdgeMoE~\cite{yi2023edgemoe} exploits sparse expert activation with hierarchical expert placement, adaptive quantization, and preloading. MobiLoRA~\cite{li2025mobilora} makes KV-cache reuse and eviction aware of both shared semantic context and mobile application state, whereas Fast On-device LLM Inference with NPUs~\cite{xu2025fast} demonstrates the need for accelerator-specific execution paths. Hybrid SLM--LLM inference~\cite{hao2024hybrid} further exposes model-capability routing as an edge--cloud systems choice; FwdLLM~\cite{xu2024fwdllm}, although focused on federated fine-tuning rather than serving, similarly shows that memory-efficient edge execution and communication-aware model adaptation are coupled for large models.

% Taken together, these studies provide essential mechanisms for heterogeneous-platform-aware inference: performance profiling, graph partitioning, memory hierarchy management, adaptive model selection, and priority-aware scheduling. However, they largely optimize a fixed DNN graph or an independent inference request. They do not jointly model the heterogeneous modules, persistent cross-step state, action timing, and robot/simulator I/O that define an embodied control loop. An embodied runtime therefore needs to retain these platform-level mechanisms while exposing module-level scheduling and state semantics above them.

\textbf{Embodied-workload-aware inference optimization.}Recent VLA policies already expose nonuniform control workloads: reusable task and semantic context is coupled with high-rate perception and action paths, often through multimodal conditioning or hierarchical planning~\cite{brohan2023rt1,rt2,driess2023palme,jiang2023vima,ahn2022saycan,rth,hirobot}. Action chunks, receding-horizon generation, autoregressive action decoding, flow-matching experts, and predictive action representations further make the control cadence stateful rather than a uniform per-step forward pass~\cite{zhao2023act,chi2023diffusionpolicy,octo,openvla,pi0,wu2024gr1,shafiullah2022bet}.

Against this background, embodied inference optimization has begun to directly redesign the VLA execution path. OpenVLA-OFT~\cite{kim2025openvlaoft} replaces autoregressive action-token generation with parallel continuous-action decoding and action chunking, while SmolVLA~\cite{shukor2025smolvla} couples a compact policy with an asynchronous inference stack. TurboVLA~\cite{xie2026turbovla} removes the LLM-centered $V\rightarrow L\rightarrow A$ path in favor of direct vision--language-to-action decoding; ActQuant~\cite{akbari2026actquant} makes mixed-precision quantization action-aware; and PolicyTrim~\cite{wang2026policytrim} increases the usable action-chunk horizon while reducing redundant physical steps. Together, these works optimize not only per-call latency and memory usage but also the number and cadence of policy invocations required to complete a task.

A second direction exploits redundancy across consecutive control steps while respecting the semantics of physical interaction. VLA-Cache~\cite{xu2025vlacache} reuses slowly changing visual-token KV states, whereas ActionCache~\cite{oi2026actioncache} retrieves intermediate actions from similar contexts to warm-start flow-matching generation. OxyGen~\cite{li2026oxygen} accelerates concurrent action and language inference through unified KV cache management across tasks and frames.
Reflex~\cite{guo2026reflex} partitions attention state into static, sliding, and dynamic regions for correct incremental cache updates, and Jetson-PI~\cite{yang2026jetsonpi} combines foresight-aligned asynchronous inference with confidence-based scheduling on onboard hardware. Deltoris~\cite{liu2026deltoris} further exploits temporal input similarity through bit-level sparsity and speculative execution. At the model-schedule level, Fast-in-Slow~\cite{fis} separates low-frequency reasoning from a high-frequency action pathway, while DAM-VLA~\cite{damvla} refreshes modality-specific latent buffers at sensor-dependent rates. These methods shift acceleration from uniform frame or token reduction to embodied-aware decisions about which state can be reused, which module must be refreshed, and when a new action chunk must be produced.

Recent WAM optimization targets both the cost of explicit future generation and the alignment of generated action chunks with reality. Fast-WAM~\cite{fastwam} retains video-based supervision during training but removes iterative future imagination from test-time control, while LaWAM~\cite{lawam} preserves predictive dynamics through compact latent visual subgoals rather than reconstructed future video. When to Trust Imagination~\cite{wang2026trustimagination} adapts the executed horizon according to predicted--observed consistency; Faster-WAM~\cite{ma2026fasterwam} decouples a lightweight action head from a deep video backbone; and MemoryWAM~\cite{yang2026memorywam} combines recent frames, event anchors, and gist tokens for persistent state. LiLa-WAM~\cite{yang2026lilawam} further moves world reasoning into a compact latent space. World Action Models in Real Time~\cite{motubrain2026realtimewam} shows experimentally that asynchronous WAM deployment requires strict temporal alignment at chunk boundaries, while Adaptive-WAM~\cite{ang2026adaptivewam} uses quality-guided exits from intermediate video-diffusion features in an autonomous-driving setting. Collectively, these results identify the world-prediction path itself as a schedulable and compressible inference component, rather than an indivisible prerequisite for control.

This recent literature is substantially closer to real embodied deployment, but each method still optimizes a particular model, state representation, or execution schedule.
What remains missing is a portable runtime that combines these useful inference optimization methods across both VLA and WAM architectures, which is the goal of \system.

\section{Project Overview}

\subsection{Challenges}
\label{subsec:challenges}

Embodied deployment should first be understood by contrast with traditional LLM and VLM inference. Conventional language or multimodal serving usually assumes a mostly synchronous request-response path, a relatively uniform token interface, and throughput-oriented optimization over large batches or many concurrent users. \system faces a different setting: embodied models run inside closed-loop control, combine multiple heterogeneous modules, and must remain deployable on robot-side hardware while absorbing new model families over time. For an open-source embodied inference runtime project, these differences collapse into three practical system challenges, each of which directly shapes the design of \system.

\begin{enumerate}
  \item \textbf{Multi-rate execution.}
    As an open-source project, \system must support not only today's embodied models, but also the optimizations that future embodied architectures may require. This is difficult because many modern embodied systems are no longer monolithic. VLA models and WAMs are increasingly assembled from multiple modules, such as perception encoders, transformer backbones, predictive branches, and action heads. Unlike a conventional LLM runtime, efficient embodied inference does not always require every module to run at every step. A perception stack may refresh less frequently, a predictive branch may run only when future estimation is needed, and an action head may need to execute at a much higher control rate.

  \item \textbf{Latency-first closed-loop control.}
    Embodied deployment also raises a harder performance problem than ordinary cloud inference. In many robotics settings, execution is effectively batch-1: a single robot or simulator must receive actions continuously, with low latency, low jitter, and predictable timing behavior. At the same time, deployment targets are heterogeneous, spanning Jetson devices, RK-based platforms, x86 edge boxes, and workstation-class systems. This creates a tension between latency-first execution and the fused-inference techniques needed to make small-batch execution efficient across different backends and devices.

  \item \textbf{Extensible embodied interfaces.}
    Finally, implementing an embodied inference runtime is not only a scheduling problem. New model families regularly introduce new dependency stacks, larger custom operators, and new input/output conventions. A practical open-source inference runtime must therefore absorb model-specific libraries, cover a broader operator surface, and support new embodied data types on both sides of the interface. Inputs may include images, language, proprioception, history, force or tactile signals, and simulator-provided state. Outputs may be discrete action tokens, continuous action vectors, action chunks, world predictions, or intermediate control representations.
\end{enumerate}

Taken together, these challenges explain why embodied deployment should not be framed as a small extension of an LLM runtime. The reusable substrate remains real and important, especially around multimodal projection, transformer-style execution, backend portability, and operator reuse. But the top-level contract has shifted from token serving to embodied control, and an open runtime must remain extensible enough to absorb future embodied optimizations rather than hard-coding today's model structure.

\subsection{Design Principles and Runtime Architecture}

These challenges, in turn, motivate three design principles of \system:
\begin{enumerate}
  \item \textbf{Modular multi-rate execution.} \system should expose explicit execution units, pluggable modules, shared state or feature pools, and configurable refresh policies so that different components can run at different rates without forcing a single synchronous path.
  \item \textbf{Latency-first fused execution.} \system should prioritize stable control performance while supporting graph replay, buffer reuse, operator fusion, backend-specific dispatch, and careful host-device data movement to enable efficient small-batch inference on heterogeneous devices.
  \item \textbf{Extensible operator and I/O support.} \system should provide typed, embodied interfaces, pluggable heads, first-class deployment adapters, and sufficient backend and operator coverage to enable new embodied paradigms to be implemented without rebuilding the surrounding runtime each time.
\end{enumerate}

The model-side analysis in Section~\ref{sec:related_work} and the runtime gap in Table~\ref{tab:runtime-compare} together motivate a runtime architecture that stays uniform at the deployment boundary while remaining extensible inside. Recent systems suggest three complementary lessons: \vlacpp{} shows that a broad family of VLA models can be organized around one shared inference path and one portable bundle format~\cite{vla-cpp}; Pie shows the value of separating a fixed execution substrate from a programmable control layer~\cite{pie}; and FlashRT shows that latency-first execution benefits from treating execution state as a first-class runtime object rather than as an implicit cache~\cite{flashrt}. \system combines these lessons in a form specialized for embodied deployment.

\begin{figure*}[t]
  \centering
  \includegraphics[width=\textwidth]{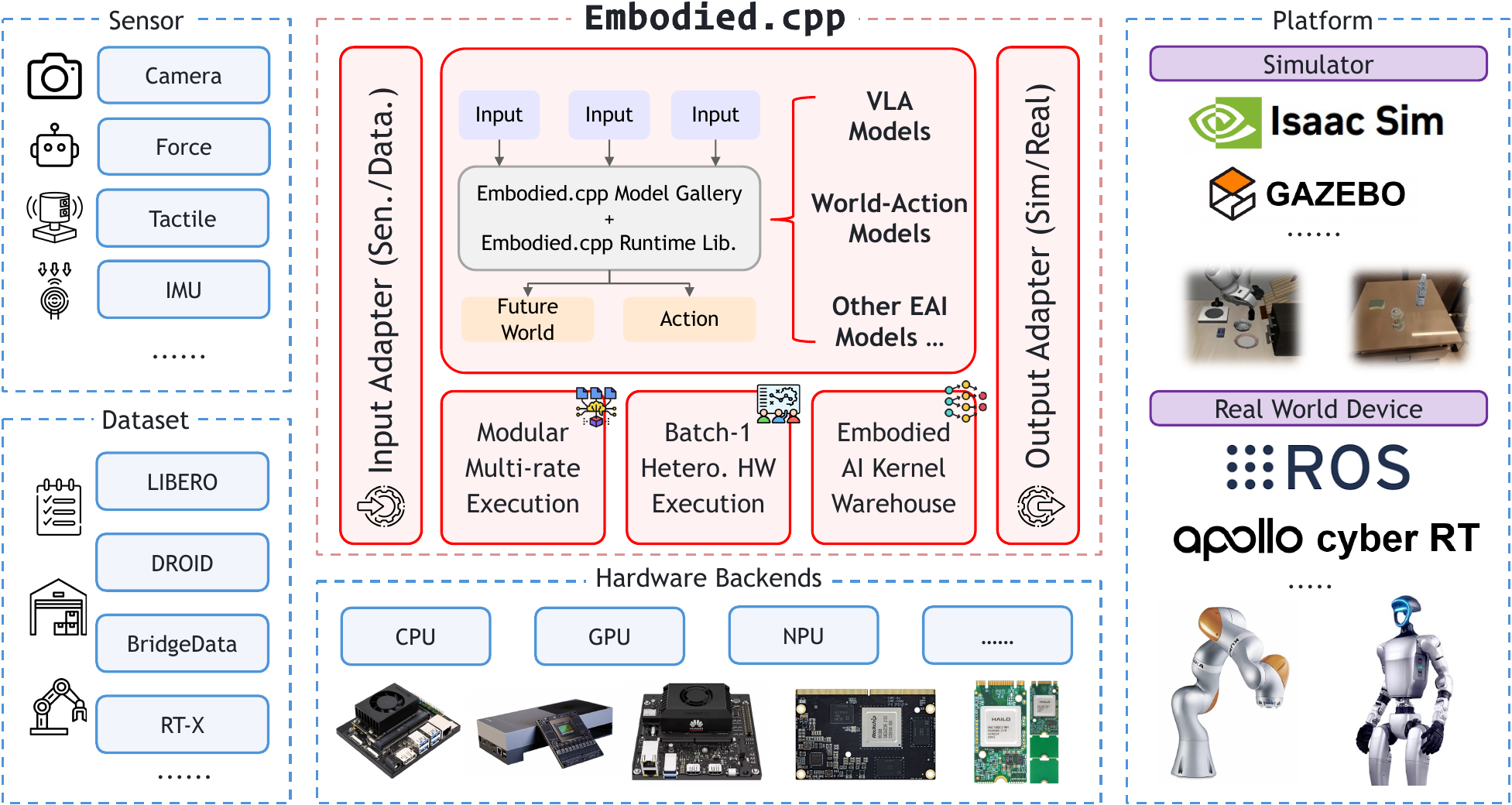}
  \caption{Project overview of \system. Diverse sensors and datasets enter through input adapters; model execution is unified around one embodied-model runtime that supports VLA models, WAMs, and future variants; and outputs are bridged to simulators and real robots through deployment adapters. Beneath this shared path, three runtime capabilities directly address the core challenges: modular multi-rate execution, latency-first batch-1 execution on heterogeneous hardware, and an embodied AI kernel warehouse for reusable operators and model-specific kernels.}
  \label{fig:system-architecture}
\end{figure*}

Figure~\ref{fig:system-architecture} summarizes this design at a high level. On the left, \emph{input adapters} absorb both online sensor streams and offline dataset samples, so cameras, force or tactile signals, IMU data, and benchmark datasets enter the runtime through one typed embodied interface. In the center, \system maintains one shared embodied-model execution zone that can host VLA models, WAMs, and future embodied variants, while preserving explicit interfaces for future prediction, action experts, and final action generation. Beneath this model zone, three supporting subsystems directly mirror the challenges in Section~\ref{subsec:challenges}: \emph{modular multi-rate execution} handles decoupled scheduling and runtime state, \emph{latency-first batch-1 execution on heterogeneous hardware} targets low-latency deployment across CPUs, GPUs, NPUs, and other accelerators, and the \emph{embodied AI kernel warehouse} collects reusable operators and model-specific kernels needed by evolving embodied architectures. On the right, \emph{output adapters} bridge runtime outputs to simulators and real-world robot software stacks. In this way, \system keeps the deployment boundary stable even as model structure, hardware targets, and embodied interfaces continue to evolve.

\section{Evaluation}

This revision reports two kinds of evidence: normalized closed-loop deployment results for pi0.5, GR00T N1.7, and HY-VLA, and deployment results for Cosmos3 and LingBot-VA. The WAM results compare Python and \system backends using success rate and VRAM, directly exposing the practical deployment trade-off.

\noindent\textbf{VLA model evaluation.} We compare Python and \system deployment paths for pi0.5, GR00T N1.7, and HY-VLA under C++, 8-bit, 6-bit, and 4-bit C++ configurations. Runtime, success rate, and VRAM are normalized independently to each model's Python baseline, which is set to $1.00$.

\begin{table*}[t]
  \centering
  \caption{Normalized VLA deployment benchmark. Values are relative to each model's Python baseline ($1.00$). Inference latency is measured per generated action step (ms/step) before normalization; lower latency and VRAM are better, while higher success rate is better.}
  \label{tab:vla_deployment_results}
  \scriptsize
  \setlength{\tabcolsep}{2.0pt}
  \renewcommand{\arraystretch}{1.14}
  \begin{tabularx}{\textwidth}{@{}>{\raggedright\arraybackslash}p{0.13\textwidth} *{15}{>{\centering\arraybackslash}X}@{}}
    \toprule
    \rowcolor{black!5}
    & \multicolumn{5}{c}{\textbf{Inference Latency}} & \multicolumn{5}{c}{\textbf{Success Rate}} & \multicolumn{5}{c}{\textbf{VRAM}} \\
    \cmidrule(lr){2-6}\cmidrule(lr){7-11}\cmidrule(lr){12-16}
    \rowcolor{black!5}
    \textbf{Model} & \textbf{Python} & \textbf{C++} & \shortstack{\textbf{C++}\\\textbf{8-bit}} & \shortstack{\textbf{C++}\\\textbf{6-bit}} & \shortstack{\textbf{C++}\\\textbf{4-bit}} & \textbf{Python} & \textbf{C++} & \shortstack{\textbf{C++}\\\textbf{8-bit}} & \shortstack{\textbf{C++}\\\textbf{6-bit}} & \shortstack{\textbf{C++}\\\textbf{4-bit}} & \textbf{Python} & \textbf{C++} & \shortstack{\textbf{C++}\\\textbf{8-bit}} & \shortstack{\textbf{C++}\\\textbf{6-bit}} & \shortstack{\textbf{C++}\\\textbf{4-bit}} \\
    \midrule
    pi0.5 & 1.00 & 0.90 & 0.88 & 0.95 & 0.88 & 1.00 & 0.92 & 0.90 & 0.93 & 0.70 & 1.00 & 0.60 & 0.41 & 0.35 & 0.30 \\
    GR00T N1.7 & 1.00 & 0.72 & 0.70 & 0.70 & 0.65 & 1.00 & 0.96 & 0.97 & 0.97 & 0.96 & 1.00 & 0.93 & 0.57 & 0.48 & 0.38 \\
    HY-VLA & 1.00 & 0.48 & 0.38 & 0.39 & 0.37 & 1.00 & 1.02 & 1.00 & 1.01 & 0.99 & 1.00 & 0.68 & 0.27 & 0.25 & 0.23 \\
    \bottomrule
  \end{tabularx}
\end{table*}

Table~\ref{tab:vla_deployment_results} preserves the Python baseline for each model and makes the deployment trade-offs comparable across model scales. The C++ paths reduce normalized inference latency for all three models, with the largest reduction for HY-VLA; 8-bit, 6-bit, and 4-bit configurations further lower VRAM relative to the corresponding Python path. Success-rate retention remains high for pi0.5 and GR00T N1.7 across the evaluated C++ configurations, while HY-VLA retains near-baseline success despite the more aggressive latency and memory reductions. Together, these results show that \system supports distinct VLA architectures and quantization levels while making their latency, memory, and control-quality trade-offs explicit.

\noindent\textbf{WAM evaluation.} We evaluate two WAM deployments, Cosmos3 and LingBot-VA, by comparing their Python implementations with the corresponding \system C++ paths. The table reports closed-loop success rate and peak VRAM for each backend.

\begin{table}[t]
  \centering
  \caption{WAM deployment results for Cosmos3 and LingBot-VA.}
  \label{tab:wam_deployment_results}
  \scriptsize
  \setlength{\tabcolsep}{1.5pt}
  \renewcommand{\arraystretch}{1.08}
  \begin{tabularx}{\columnwidth}{@{}>{\raggedright\arraybackslash}p{0.30\columnwidth} >{\centering\arraybackslash}p{0.18\columnwidth} >{\centering\arraybackslash}p{0.23\columnwidth} >{\centering\arraybackslash}X@{}}
    \toprule
    \rowcolor{black!5}
    \textbf{Model} & \textbf{Backend} & \textbf{Success Rate} & \textbf{VRAM} \\
    \midrule
    Cosmos3 & Python & 49.00\% & 21.84~GB \\
    Cosmos3 & C++ & 48.00\% & 19.49~GB \\
    LingBot-VA & Python & 100.00\% & 24.75~GB \\
    LingBot-VA & C++ & 98.00\% & 16.44~GB \\
    \bottomrule
  \end{tabularx}
\end{table}

Table~\ref{tab:wam_deployment_results} shows that the C++ paths preserve nearly all task success while reducing memory. Cosmos3 retains 48.00\% success versus 49.00\% for Python and lowers VRAM from 21.84~GB to 19.49~GB. LingBot-VA retains 98.00\% success versus 100.00\% and lowers VRAM from 24.75~GB to 16.44~GB, a 33.6\% reduction. These results provide direct evidence that \system can deploy distinct WAM architectures with limited task-quality loss and substantially lower memory demand.

%\section{Conclusion}
%
%Embodied models are becoming more diverse, but deployment infrastructure remains comparatively underdeveloped. \system addresses this gap by framing embodied deployment as a portable C++ runtime problem rather than as a collection of model-specific integration scripts. The resulting design emphasizes unified execution, backend portability, operator extensibility, and robot-side integration across heterogeneous edge devices and robots.
%
%As embodied paradigms continue to evolve, the need for a stable deployment substrate will only become more important. \system is intended to provide that substrate by making embodied models easier to run, compare, and integrate in practical edge and robot settings.

\section{Conclusion}

The evidence from recent models points in one direction: embodied deployment is converging on a shared execution path even though model families continue to diversify. \system captures this convergence in an inference runtime that treats the common path as infrastructure and the diverging parts as plugins. Its five-layer architecture keeps interaction pattern, I/O semantics, objective, and deployment boundary explicit, while reusing the same backbone execution path across paradigms. In the current revision, we validate the C++ inference path quantitatively on three VLA models and position WAMs through architectural analysis. As new embodied model variants emerge, this separation between a stable core and pluggable task-specific components will only compound in value.

\bibliographystyle{unsrturl}
\bibliography{ref}

\end{document}